%% file: root.tex
\documentclass[letterpaper, 10 pt, conference]{ieeeconf}  

\IEEEoverridecommandlockouts                              

\usepackage{graphics} 
\usepackage{epsfig} 
\usepackage{mathptmx} 
\usepackage{times} 
\usepackage{amsmath} 
\usepackage{amssymb}  
\usepackage{booktabs}
\usepackage{url}
\usepackage{xcolor}

\usepackage{hyperref}

\title{\LARGE \bf
IMPACT-Scribe: Interactive Temporal Action Segmentation with Boundary Scribbles and Query Planning
}

\author{Qian Yin$^{1\dagger}$, Di Wen$^{1\dagger*}$, Kunyu Peng$^{1,2}$, David Schneider$^{1}$, Zeyun Zhong$^{1}$, Alexander Jaus$^{1}$, Zdravko Marinov$^{1}$,\\ Jiale Wei$^{1}$, Ruiping Liu$^{1}$, Junwei Zheng$^{1,3}$, Yufan Chen$^{1}$, Chen Zhang$^{1}$, Lei Qi$^{4}$, and Rainer Stiefelhagen$^{1}$%
\thanks{$^{\dagger}$These authors contributed equally to this work.}%
\thanks{$^{*}$Corresponding author: Di Wen ({\tt\small di.wen@kit.edu}).}%
\thanks{$^{1}$The authors are with Karlsruhe Institute of Technology, 76131 Karlsruhe, Germany.}%
\thanks{$^{2}$The author is with INSAIT, Sofia University “St. Kliment Ohridski”, 1784 Sofia, Bulgaria.}%
\thanks{$^{3}$The author is with ETH Zurich, 8092 Zurich, Switzerland.}%
\thanks{$^{4}$The author is with Technical University of Munich, 80333 Munich, Germany.}%
}

\begin{document}

\maketitle
\thispagestyle{empty}
\pagestyle{empty}



\input{sec/abstract}
\input{sec/introduction}
\input{sec/related}

\input{sec/methodology}

\input{sec/experiment}
\input{sec/conclusion}




\bibliographystyle{IEEEtran}
\bibliography{main}
\addtolength{\textheight}{-12cm}   

\end{document}

%% file: sec/abstract.tex
\begin{abstract}
Dense temporal annotation of procedural activity videos is vital for action understanding and embodied intelligence but remains labor-intensive due to reactive tools. Each correction is treated as an isolated edit, limiting reuse of information on annotator uncertainty and model reliability. We introduce \textbf{IMPACT-Scribe}, a correction-driven framework for dense labeling that uses each correction to improve future human--machine collaboration. IMPACT-Scribe combines uncertainty-aware boundary scribble supervision, local proposal modeling, cost-aware query planning, structured propagation, and correction-driven adaptation. Experiments and a human study show that this closed-loop design improves labeling quality per effort, enhances boundary accuracy, and fosters better human-machine interaction over time.
The code will be made publicly available at
\href{https://github.com/BanzQians/IMPACT_AS}{https://github.com/BanzQians/IMPACT\_AS}.
\end{abstract}

%% file: sec/introduction.tex
\section{Introduction}

The demand for large-scale, densely labeled video data has surged with advances in temporal action understanding~\cite{farha2019ms,yi2021asformer} and embodied intelligence~\cite{li2026egocross}. However, the human effort required for labeling has not kept pace. For procedural activity videos, annotation cost arises not from isolated decisions but from the cumulative burden of repeated local corrections under uncertainty across long, visually homogeneous, and temporally ambiguous sequences. Thus, the core bottleneck is not only a learning issue but also a human--machine collaboration challenge.
Temporal action segmentation~\cite{yi2021asformer,peng2025hopadiff} presents several challenges that amplify the annotation burden. First, action boundaries in procedural videos are often gradual, making boundary placement inherently uncertain. Second, neighboring actions may be highly similar, requiring annotators to consider extended temporal context instead of isolated frames. Third, errors accumulate over long sequences, leading to fatigue and reduced consistency. These challenges make dense temporal annotation interaction-intensive, yet most existing tools offer limited uncertainty-aware assistance at the sequence level.
Prior work has attempted to reduce annotation effort in temporal
action segmentation through weak supervision, including timestamp
labels~\cite{li2021temporal,rahaman2022generalized}, set-level
constraints~\cite{lu2022set}, and active frame
selection~\cite{su2024two}. These methods reduce the amount of
supervision required, but they generally treat human input as a
fixed offline signal rather than as part of an evolving
interactive process. Interactive video annotation
systems~\cite{oh2019fast,heo2021guided,yin2021learning} have
shown that local user corrections can be propagated across a
sequence, but they still largely handle each correction as an
isolated edit. In particular, they make limited use of what a
correction reveals about annotator uncertainty and local model
reliability; they also do not fully exploit the information such
corrections provide about likely future failure regions or the
effort associated with different forms of intervention. As a
result, the interaction loop remains largely reactive: the system
helps resolve the current mistake, but does not substantially
improve how it supports the next one.

In this paper, we adopt a different perspective, arguing that each human correction in dense temporal annotation should not only be a local edit but also serve as supervision for how the system should collaborate with the annotator. This turns annotation into a closed-loop human-in-the-loop process, where each interaction refines the current labeling, informs structured propagation, and improves future system behavior. This approach is particularly crucial for procedural videos, where annotation costs are driven by repetitive, locally ambiguous corrections, and the process's efficiency hinges on how well the system prioritizes, interprets, and reuses human feedback.

Motivated by this perspective, we present \textbf{IMPACT-Scribe},
a correction-driven interactive framework for dense temporal action
annotation. Rather than forcing annotators to commit to frame-exact
boundaries, \textbf{IMPACT-Scribe} encodes temporal ambiguity
through scribble-based supervision. Each scribble is mapped by a
lightweight local proposal model into a structured local
correction, which is then integrated into a globally coherent
sequence labeling through constrained propagation. At the
interaction level, cost-aware query planning allocates human
effort based on expected sequence-level benefit rather than
prediction uncertainty alone. Crucially, correction-driven
adaptation continuously updates query utility, effort estimation,
confidence calibration, and proposal behavior from accumulated
correction history, forming a closed-loop system in which
interaction, inference, and adaptation are tightly coupled and
progressively improve over time.

Our contributions are threefold.
\begin{itemize}
\item We frame dense temporal action annotation as a \emph{correction-driven human-in-the-loop collaboration problem}, where each correction refines labeling and improves future system behavior.

\item We introduce \textbf{IMPACT-Scribe}, an uncertainty-aware annotation framework combining scribble-based supervision, proposal generation, cost-aware query planning, and dense propagation for efficient sequence-level corrections.

\item We develop correction-driven adaptation mechanisms that update query utility, interaction cost, confidence, and proposal behavior from past corrections, demonstrating through experiments that this design reduces annotation burden while maintaining labeling quality.
\end{itemize}

%% file: sec/related.tex
\section{Related Work}

\subsection{Temporal Action Segmentation}
The central challenge in Temporal Action Segmentation (TAS) is not just prediction accuracy but the cost of obtaining the dense per-frame supervision that accurate models require. Early architectures addressed this through multi-stage temporal convolutions refining over-segmented predictions~\cite{lea2017tcn,farha2019ms,li2020mstcnpp}, while transformer-based designs extended temporal receptive fields while preserving local boundary sensitivity~\cite{yi2021asformer,lu2024fact}. Annotation costs have been reduced by weakening supervision: timestamp labels minimize per-video cost~\cite{li2021temporal,rahaman2022generalized,wang2026tqt}, set-level constraints eliminate ordering requirements~\cite{lu2022set}, active selection focuses labeling on informative samples~\cite{su2024two}, and few-shot or cross-view transfer extend learned representations to new domains with minimal annotation~\cite{lu2025mmftas,quattrocchi2026exoego}. However, these approaches reduce the \emph{quantity} of human input but do not address its \emph{quality} as an interactive signal. In all cases, human input is treated as a fixed, single-pass observation, consumed by the model and discarded, rather than an ongoing source of information about annotator uncertainty, model failure, and evolving annotation costs.
\subsection{Interactive and Active Annotation}
Interactive annotation research has evolved from focusing on \emph{what} can be corrected to \emph{how efficiently} corrections can be allocated. In spatial segmentation, this shift moved from backpropagation-based refinement~\cite{jang2019brs} to feature-space updates~\cite{sofiiuk2020fbrs}, and finally to architectures that limit inference to the local neighborhood of each click~\cite{chen2022focalclick,liu2023simpleclick}. In video annotation, this trajectory progressed from interaction-and-propagation pipelines~\cite{oh2019fast} to reliability-guided recommendations~\cite{heo2021guided}, and policies that optimize frame selection and annotation type for mask improvement~\cite{yin2021learning,qiao2023hvsa}. Active annotation further closes the loop by asking where to look next, using uncertainty-diversity samplers~\cite{heilbron2018annotate}, annotation-type allocation~\cite{rana2025omvid}, and gain-per-cost ranking~\cite{gwon2025candidate}. Realistic evaluation requires faithful annotator simulation~\cite{marinov2024rethinking}, motivating our oracle-controlled offline evaluation. However, this body of work does not address dense temporal boundary annotation, where corrections are gradual, errors compound over frames, and each edit provides latent information about system reliability and future annotation cost.

\subsection{Procedural Structure and Action-Centric Extensions}
Understanding procedural activity requires reasoning beyond individual frame predictions. Segmental grammars and temporal logic integrated duration and ordering constraints~\cite{pirsiavash2014parsing,xu2022dtl}, while task-graph formulations extended this to real-time segmentation and error detection~\cite{seminara2024tgml,shen2024protas}. The IMPACT benchmark~\cite{wen2026impact} addresses industrial assembly tasks with long-horizon actions, occlusion, and multi-route workflows, where annotation cost is high and current tools fall short. Work on label noise and domain adaptation~\cite{xu2024noiseerasar,fan2025drivernoisy} emphasizes that real-world annotations are noisy and dynamic. IMPACT-Scribe tackles this by improving the annotation process, treating each correction as a signal that updates labeling, guides propagation, and supervises future human--machine collaboration.

%% file: sec/methodology.tex
\section{Methodology}
\label{sec:method}

\begin{figure*}[t]
    \centering
    \includegraphics[width=\textwidth]{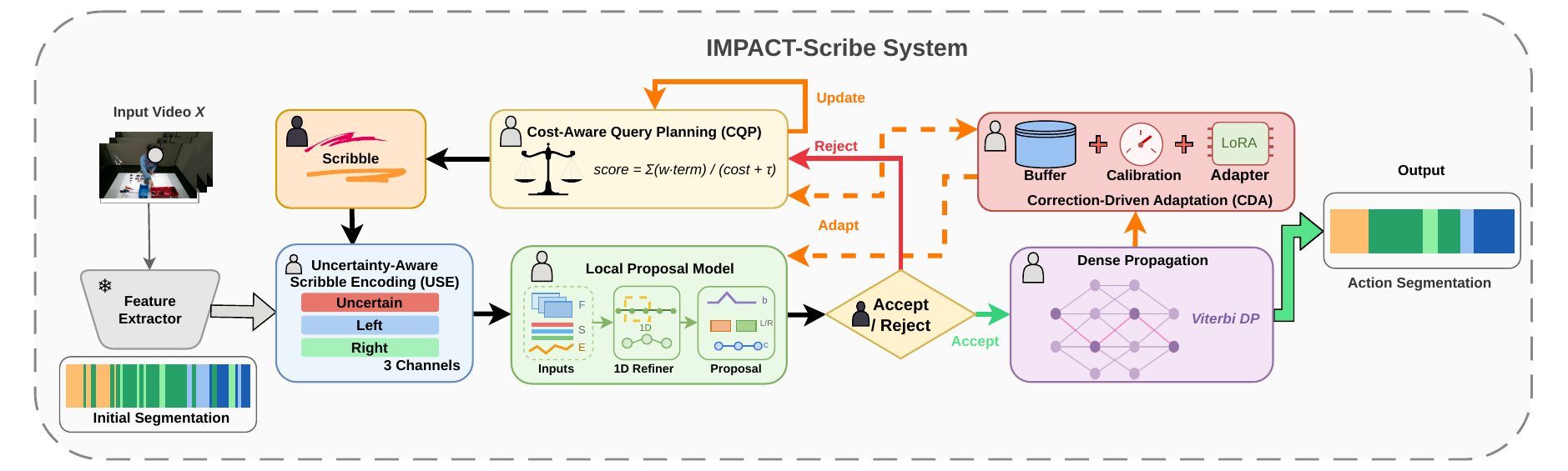}
    \vskip-3ex
    \caption{
        \textbf{Overview of the IMPACT-Scribe system.}
        Given an input video, a frozen feature extractor produces dense embeddings that support five core components.
        \emph{Uncertainty-Aware Scribble Encoding (USE, \S\ref{subsec:use})} converts annotator scribbles into a 3-channel temporal signal (uncertain / left / right).
        The \emph{Local Proposal Model (\S\ref{subsec:local_model})} predicts boundary corrections from the encoded scribble and dense features.
        \emph{Cost-Aware Query Planning (CQP, \S\ref{subsec:cqp})} ranks candidate queries using a utility--cost tradeoff and selects the next one for user validation.
        \emph{Dense Propagation (\S\ref{subsec:propagation})} applies Viterbi DP to propagate accepted corrections into a globally coherent action segmentation.
        \emph{Correction-Driven Adaptation (CDA, \S\ref{subsec:cda})} updates the planner and proposal model from accepted/rejected feedback, enabling continual human-in-the-loop adaptation.
    }
    \label{fig:impact_scribe_overview}
    \vskip-5ex
\end{figure*}

IMPACT-Scribe is a correction-driven human-in-the-loop framework for dense temporal annotation. As illustrated in Figure~\ref{fig:impact_scribe_overview}, it consists of five coupled components: \emph{Uncertainty-Aware Scribble Encoding (USE)}, a \emph{Local Proposal Model}, \emph{Cost-Aware Query Planning (CQP)}, \emph{Dense Propagation}, and \emph{Correction-Driven Adaptation (CDA)}. USE, CQP, and CDA define the core interaction logic of the system: USE converts annotator uncertainty into a structured temporal input, CQP allocates attention to queries expected to provide the largest benefit relative to effort, and CDA reuses confirmed interaction outcomes to improve later assistance. The Local Proposal Model is the main learnable module, while Dense Propagation enforces sequence-level coherence through structured inference rather than a separate end-to-end dense predictor.

\subsection{Problem Formulation}
\label{subsec:problem}

Let $X=\{x_t\}_{t=1}^{T}$ denote a video sequence, let $F=\{f_t\}_{t=1}^{T}$ be precomputed temporal features extracted by an off-the-shelf I3D~\cite{carreira2017quo}, and let $\mathcal{L}$ be the action label vocabulary. The goal is to recover a dense frame-level labeling $Y=\{y_t\}_{t=1}^{T}$, where $y_t\in\mathcal{L}$, while reducing the amount of human effort required to obtain it.


At interaction step $k$, the system maintains a current sequence hypothesis $Y^{(k)}$, selects a query $q^{(k)}$, receives an annotator scribble $S^{(k)}$, maps it to a structured local correction $A^{(k)}$ through the local proposal model $M_\theta$, and propagates the accepted correction into an updated dense labeling:
\begin{equation}
Y^{(k)} \xrightarrow{\text{plan}} q^{(k)}
\xrightarrow{\text{annotate}} S^{(k)}
\xrightarrow{M_\theta} A^{(k)}
\xrightarrow{\text{propagate}} Y^{(k+1)}.
\label{eq:loop}
\end{equation}
When an external initialization $Y_{\mathrm{init}}$ is available, we set $Y^{(0)}=Y_{\mathrm{init}}$; otherwise the system starts from an empty annotation state. Here, $M_\theta$ denotes the local proposal model introduced in \S\ref{subsec:local_model}. Each accepted correction additionally updates the utility estimator, effort model, confidence calibration, and correction memories used in subsequent interactions. IMPACT-Scribe therefore treats correction not only as an edit to the current sequence, but also as feedback for improving later collaboration.

\subsection{Uncertainty-Aware Scribble Encoding (USE)}
\label{subsec:use}
Action boundaries in procedural video are often gradual rather than frame-exact, so we represent corrective input with an \emph{uncertain temporal boundary scribble}, encoding both a candidate boundary location and an ambiguity range. This explicitly preserves annotator uncertainty instead of forcing a single-frame decision.

The same scribble primitive supports multiple correction intents via temporal extent and gesture semantics. Broad horizontal strokes indicate uncertain boundary localization, vertical strokes signal edit cues (e.g., deletion or refinement), and a single stroke spanning multiple boundaries triggers merged corrections over a wider temporal area. Thus, USE functions as both an input representation and a lightweight interaction mechanism.


For each interaction, we define a local temporal window $W=[a,b]\subseteq[1,T]$ by expanding the scribble support on both sides with a bounded context radius and clipping the result to the valid video range. This provides sufficient temporal context around the uncertain boundary while keeping inference local. The scribble is projected from the interface canvas onto the temporal axis and converted into normalized support channels over $W$. These channels encode uncertain support together with optional one-sided refinement evidence, and they induce an uncertain interval $I^{+}(S)\subseteq[1,T]$ used in both inference and training. The output of USE is therefore a structured local interaction state, consisting of the support channels and the induced interval constraint, which is passed directly to the local proposal model.

\subsection{Local Proposal Model}
\label{subsec:local_model}

Given local temporal features $F(W)=\{f_t\}_{t\in W}$, the USE-derived scribble representation $S(W)$, and a normalized boundary-energy signal $E(W)$ computed over the same interaction window, the local proposal model $M_\theta$ predicts a structured local update:
\begin{equation}
p_b(t\mid W,S),\quad p_L(y\mid W,S),\quad p_R(y\mid W,S),\quad c(W,S),
\label{eq:outputs}
\end{equation}
where $y\in\mathcal{L}$. These outputs correspond to a boundary posterior, left/right side-label posteriors, and a proposal confidence score.

As shown in Figure~\ref{fig:impact_scribe_overview}, the module consists of three stages: input assembly from $F(W)$, $S(W)$, and $E(W)$; a lightweight 1D temporal refiner; and output heads for boundary localization, side-label prediction, and confidence estimation. Inference is restricted to a compact local window centered on the interaction, which reflects the locality of most boundary errors and helps keep response latency low.

This module is the primary trainable component of IMPACT-Scribe. Its role is not to predict dense labels from scratch, but to interpret uncertainty-aware human input as a structured local proposal that can be accepted directly or further propagated downstream. The learning target is therefore \emph{correction completion under human guidance}, rather than unconstrained temporal classification.

To improve label assistance over time, the system also maintains lightweight memories derived from accepted corrections. In particular, it stores prototype statistics for previously accepted segments and confusion statistics extracted from correction history. These memories do not replace the proposal model; instead, they bias candidate suggestions toward labels that are both locally plausible and historically consistent with prior corrections.

\subsection{Cost-Aware Query Planning (CQP)}
\label{subsec:cqp}

A candidate query is defined as $q=(t_q,W_q)$, where $t_q$ is a candidate boundary location and $W_q$ is its associated local temporal window, centered on $t_q$ and defined in the same way as the local window $W$ used by USE. Not all unresolved candidate boundaries are equally worth presenting to the annotator. IMPACT-Scribe therefore ranks candidate boundary queries by expected benefit per unit interaction cost:
\begin{equation}
\Pi(q)=\frac{U(q)}{\hat{C}(q)+\tau},
\label{eq:priority}
\end{equation}
where $U(q)$ estimates the expected value of correcting query $q$, $\hat{C}(q)$ estimates the annotator effort required for that interaction, and $\tau>0$ stabilizes the score when estimated cost is small.


This ratio-form objective reflects the budgeted nature of interactive annotation: a useful query is not just uncertain but also worth the human effort to resolve. Accordingly, $U(q)$ aggregates signals related to annotation value, including boundary ambiguity, label disagreement, expected propagation gain, and correction history, making query selection \emph{value-aware}. The scoring structure remains fixed for stability, while utility and cost estimates are updated from observed corrections, allowing the planner to progressively calibrate to the annotator and session dynamics over time.
\subsection{Dense Propagation}
\label{subsec:propagation}

Once a local correction is accepted, its effect should extend beyond a single timestamp. Dense Propagation therefore inserts the correction as an anchor constraint and updates the dense labeling through structured inference rather than by re-running a full-sequence predictor.

Let an accepted local correction at interaction step $k$ be written as
\begin{equation}
A^{(k)}=(s^{(k)},e^{(k)},b^{(k)},y_L^{(k)},y_R^{(k)}),
\label{eq:anchor}
\end{equation}
where $[s^{(k)},e^{(k)}]$ is the affected span, $b^{(k)}$ is the accepted boundary, and $y_L^{(k)},y_R^{(k)}$ are the confirmed side labels. Given the current sequence hypothesis $Y^{(k)}$ and the set of $K\leq k$ accepted anchors $\{A^{(j)}\}_{j=1}^{K}$ accumulated up to the current interaction step, propagation computes the next sequence labeling as
\begin{equation}
Y^{(k+1)}=
\arg\max_{Y'}
\sum_{t=1}^{T} \psi_t(y'_t)
-\sum_{t=2}^{T}\gamma_t\,\mathbf{1}[y'_t\neq y'_{t-1}]
+\sum_{j=1}^{K} \phi_j(Y';A^{(j)}),
\label{eq:decode}
\end{equation}
where $\psi_t(y)$ denotes local emission preferences, initialized from the cached sequence hypothesis and adjusted by the side-label posteriors of the local proposal model in the affected window; $\gamma_t$ is a frame-dependent transition penalty derived from confidence in the current sequence hypothesis, with larger values in confident regions to suppress unnecessary label switches; and $\phi_j(Y';A^{(j)})$ is a high-weight soft anchor-consistency term encouraging dense labelings that remain consistent with the accepted correction $A^{(j)}$.

The decoding objective enforces consistency at the sequence level, while the accepted cut location $b^{(k)}$ is additionally protected during writeback so that confirmed boundaries are preserved in the updated representation. This module is intentionally formulated as structured inference rather than as another learned predictor: the local proposal model interprets uncertain human input, and dense propagation ensures that accepted local edits are integrated into a globally coherent sequence.

\subsection{Correction-Driven Adaptation (CDA)}
\label{subsec:cda}

A central property of IMPACT-Scribe is that correction history is reused to improve later interactions. After each accepted correction, CDA updates lightweight statistics for query utility, effort estimation, confidence calibration, and label-confusion memory. Because these updates operate on low-dimensional summaries rather than full dense re-inference, they add negligible latency to the live interaction loop.

Adaptation proceeds at two timescales. At the fast timescale, updates are applied immediately after each accepted interaction, so that subsequent query ranking, effort estimation, and confidence calibration already reflect the latest correction outcome. At the slow timescale, high-confidence accepted corrections are buffered and used for periodic background refinement of the local proposal model once a sufficient batch of new accepted examples has been collected. Refinement is performed on a checkpointed copy rather than on the live interactive instance, which allows the framework to improve over time without destabilizing real-time use.

CDA thus provides a second adaptive layer on top of the local proposal mechanism. Rather than relying on end-to-end online retraining, it selectively updates the components most closely tied to collaboration quality. In this sense, it operationalizes the central premise of IMPACT-Scribe: each correction is both an edit to the current annotation and supervision for improving future human--machine interaction.

\subsection{Training Objective}
\label{subsec:training}

In this subsection, $S$ denotes a generic training scribble corresponding to the interaction-time scribble $S^{(k)}$ in Eq.~\eqref{eq:loop}. During training, scribbles are synthesized from ground-truth segmentations by perturbing boundary locations with random temporal offsets and interval widths, with optional left/right support strokes added in neighboring segments. The local proposal model is trained using interval-based scribble supervision rather than frame-exact point targets.

The same local window $W=[a,b]$ defined in \S\ref{subsec:use} is used during training. Given the uncertain interval $I^{+}(S)$ induced by the current scribble under USE, boundary supervision marginalizes probability mass over that interval:
\begin{equation}
\mathcal{L}_{\mathrm{boundary}}=
-\log\sum_{t\in I^{+}(S)}p_b(t\mid W,S).
\label{eq:loss_boundary}
\end{equation}
This objective assigns credit to any boundary prediction that falls within the annotator-specified ambiguity range, which better matches the semantics of uncertain correction input than single-frame cross-entropy.

Side-label supervision is defined as
\begin{equation}
\mathcal{L}_{\mathrm{side}}=
\mathrm{CE}(p_L,y_L)+\mathrm{CE}(p_R,y_R),
\label{eq:loss_side}
\end{equation}
where $y_L,y_R\in\mathcal{L}$ are the ground-truth labels on the left and right side of the target boundary, respectively. A consistency regularizer is further imposed to penalize predictions that violate already accepted local structure, including anchor inconsistency and side-label disagreement around confirmed corrections:
\begin{equation}
\mathcal{L}_{\mathrm{consistency}}=
\frac{1}{|\mathcal{A}(W,S)|}\sum_{\varphi\in\mathcal{A}(W,S)}\varphi\!\left(p_b,p_L,p_R\right),
\label{eq:loss_consistency}
\end{equation}
where $\mathcal{A}(W,S)$ denotes the set of activated consistency penalties for the current local training instance defined by window $W$ and scribble $S$.

The full training objective is
\begin{equation}
\mathcal{L}_{\mathrm{local}}=
\lambda_b\,\mathcal{L}_{\mathrm{boundary}}
+\lambda_s\,\mathcal{L}_{\mathrm{side}}
+\lambda_c\,\mathcal{L}_{\mathrm{consistency}}.
\label{eq:loss}
\end{equation}
In our implementation, we use fixed loss weights $(\lambda_b,\lambda_s,\lambda_c)=(1.0,1.0,0.3)$. These values were adopted as default settings from synthetic-scribble pretraining and held constant during deployment and online adaptation. The two direct supervised terms carry unit weight, while the smaller consistency weight regularizes local structure without overwhelming the direct scribble supervision.

In practice, the local proposal model is initialized through synthetic-scribble pretraining and subsequently refined from accepted corrections accumulated during interaction. This training protocol aligns the learnable core of IMPACT-Scribe with the correction process it is intended to support.

%% file: sec/experiment.tex
\section{Evaluation}
\label{sec:evaluation}

We evaluate IMPACT-Scribe as an interactive annotation system, focusing
not only on final dense-label quality but also on how effectively the
correction loop converts human effort into useful supervision. The
evaluation addresses four questions: 1) Does full assistance improve
annotation quality under live human interaction? 2) Does it improve
user behavior and perceived workload? 3) Does controlled offline
analysis support the mechanism-level claims? 4) Is the system
sufficiently responsive for practical deployment?

\subsection{Evaluation Protocol}
\label{subsec:eval_protocol}

\noindent\textbf{Datasets and Features.}
We use \textbf{IMPACT} as the primary benchmark and
\textbf{EPIC-KITCHENS} as a supplementary cross-scene setting for
user-study validation. IMPACT includes 13 subjects, 112 recordings, and
560 cases (one per camera viewpoint), split into 8/2/3 subjects for
training, validation, and testing. Videos are represented by offline
cached I3D Inception-RGB features~\cite{carreira2017quo}.

\noindent\textbf{Evaluation Settings.}
We report results under two complementary settings. First, a
\emph{live within-subject user study} evaluates end-to-end usability,
workload, and final annotation quality under real human interaction.
Second, a \emph{controlled offline evaluation} on the IMPACT held-out
test split (135 cases) is used for matched-budget policy comparison
and component analysis: each system variant starts from the same
prelabel initialization, runs under the same per-case interaction
budget $\lceil 1.5 \times |\mathcal{B}_{\text{GT}}|\rceil$, and uses a
GT-based oracle as the answerer. These oracle-controlled scores should
be interpreted as upper-bound process diagnostics rather than as
user-facing end-to-end performance. In the controlled offline evaluation, one interaction step denotes one oracle-accepted boundary correction. In the live user study, we distinguish accepted events from total interaction steps: accepted events count accepted boundary corrections, whereas total interaction steps count all logged user interaction actions during a session.

\noindent\textbf{Compared Conditions.}
We consider three conditions. \textbf{VIA} is a manual baseline with
no system assistance. \textbf{Scribble Only} uses the same local
correction interface but disables planner-driven assistance and CDA.
\textbf{Full Assist} is the complete IMPACT-Scribe system. Component-level
ablations are reported in \S\ref{subsec:eval_mechanism}.

\noindent\textbf{Metrics.}
\emph{Annotation-quality metrics} include Boundary F1@$\delta$ for
$\delta\in\{5,10,25,50\}$ and Edit Score. \emph{Interaction-efficiency metrics} include accepted-event counts and total interaction steps per session in the live study, as well as step- and time-budget curves in the controlled offline evaluation. \emph{Subjective metrics} are the post-task Single-Ease Question (SEQ; 1--7 Likert) and the NASA-TLX workload questionnaire.

\noindent\textbf{Implementation Details.}
All results use the default configuration from \S\ref{sec:method}
unless stated otherwise.

\subsection{Live Human Study on IMPACT and EPIC-KITCHENS}
\label{subsec:eval_human}


Eleven participants from informatics, mechatronics, or computer vision backgrounds, all with limited annotation experience, took part after informed consent; exclusions were limited to session-level data loss.

Each participant completed a short warm-up on a dedicated tutorial clip
and then annotated three IMPACT clips, one per condition (mean duration
$\approx 30$\,s, $\approx 7$ ground-truth segments, matched in length,
segment count, and label vocabulary). Condition order was fully
counterbalanced; each participant annotated one clip per condition and
no study clip was reused in warm-up. Mean task duration was
$328{\pm}210$\,s for VIA, $210{\pm}110$\,s for \emph{Scribble Only},
and $277{\pm}196$\,s for \emph{Full Assist} ($n{=}11$).

Table~\ref{tab:main_summary} reports an \emph{assisted-only diagnostic
comparison} of interaction behavior between \emph{Scribble Only} and
\emph{Full Assist}. Because the two conditions share the same
scribble-based proposal-and-confirmation interface once a correction
region is selected, this comparison isolates how full assistance
reshapes the interaction trajectory beyond the local interaction
primitive. Boundary-quality differences between the two conditions are
reported alongside VIA in Table~\ref{tab:human_main}; the paired
Wilcoxon contrast for F1@5 between the two assisted conditions reaches
$p{=}0.031$.

\begin{table}[t]
    \centering
    \caption{
        \textbf{Interaction-behavior diagnostic from the live IMPACT
        user study} ($n{=}11$, within-subject, assisted conditions
        only).
    }
    \label{tab:main_summary}
    \setlength{\tabcolsep}{6pt}
    \renewcommand{\arraystretch}{1.10}
    \begin{tabular}{lcc}
        \toprule
        \textbf{Condition}
        & \textbf{Accepted Events $\uparrow$}
        & \textbf{Total Steps $\downarrow$} \\
        \midrule
        Scribble Only & 13.09 & \textbf{30.64} \\
        Full Assist   & \textbf{21.82} & 49.36 \\
        \bottomrule
    \end{tabular}
    \vskip-4ex
\end{table}

Relative to \emph{Scribble Only}, \emph{Full Assist} elicits more accepted boundary events ($p{=}.050$) at the cost of more total steps ($p{=}.054$), based on paired two-sided Wilcoxon signed-rank tests. This indicates that the full system does not simply reduce interaction volume; rather, it reallocates effort toward more fine-grained and iterative interventions that are most beneficial when precise temporal placement matters.

Table~\ref{tab:human_main} summarizes the main user-facing outcomes
across all three conditions, including VIA. 

\begin{table}[t]
    \centering
    \caption{
        \textbf{Main live human-study outcomes on IMPACT.}
        Median [Q1,\,Q3] for time and subjective ratings;
        final saved-annotation quality for boundary metrics.
    }
    \label{tab:human_main}
    \setlength{\tabcolsep}{3.6pt}
    \renewcommand{\arraystretch}{1.10}
    \resizebox{\columnwidth}{!}{%
    \begin{tabular}{lcccccc}
        \toprule
        \textbf{Condition}
        & \textbf{Time (s) $\downarrow$}
        & \textbf{SEQ $\uparrow$}
        & \textbf{Workload $\downarrow$}
        & \textbf{F1@5 $\uparrow$}
        & \textbf{F1@10 $\uparrow$}
        & \textbf{F1@25 $\uparrow$} \\
        \midrule
        VIA
            & 217.26 [175.11, 386.49]
            & 4.0 [3.0, 6.0]
            & 3.17 [2.17, 4.08]
            & 0.134 & 0.316 & 0.421 \\
        Scribble Only
            & \textbf{207.12} [129.12, 298.92]
            & \textbf{6.0} [5.5, 6.0]
            & \textbf{2.17} [1.58, 3.25]
            & 0.123 & 0.344 & \textbf{0.642} \\
        Full Assist
            & 255.07 [142.44, 323.24]
            & 5.0 [4.5, 6.0]
            & 2.50 [1.75, 3.08]
            & \textbf{0.257} & \textbf{0.386} & 0.635 \\
        \bottomrule
    \end{tabular}%
    }
\end{table}

Both assisted interfaces outperform manual VIA at moderate tolerances and reduce perceived workload. \emph{Full Assist} yields the strongest strict-tolerance F1; \emph{Scribble Only} is descriptively faster, simpler, and slightly stronger at looser tolerances.

\begin{figure}[t]
    \centering
    \vskip-2ex
    \includegraphics[width=0.45\columnwidth]{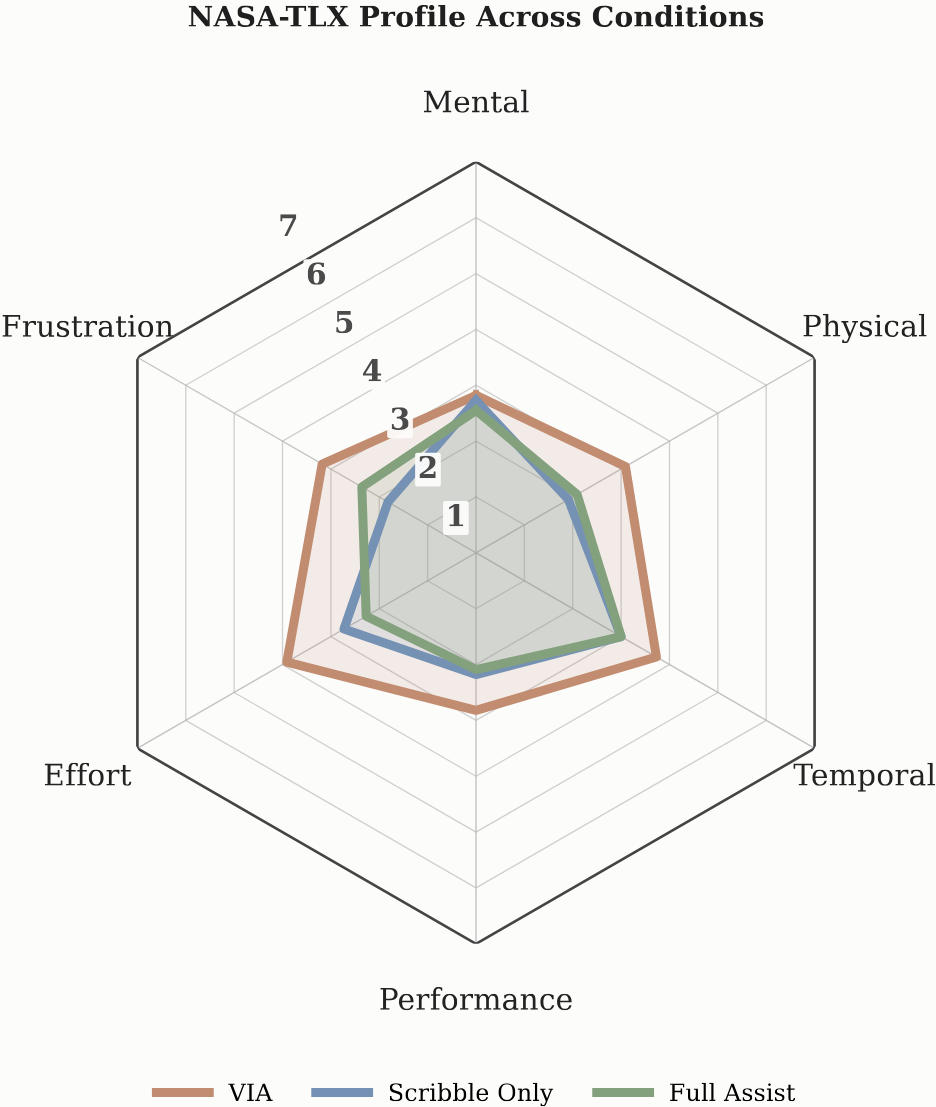}
    \vskip-1ex
    \caption{\textbf{NASA-TLX workload profile on IMPACT} ($\downarrow$
    on all axes).}
    \label{fig:nasa_tlx_radar}
    \vskip-2ex
\end{figure}

Figure~\ref{fig:nasa_tlx_radar} summarizes the NASA-TLX workload
profile; significant effects are observed for perceived effort and
physical demand, with both assisted conditions rated lower than VIA.

Similar trends were observed on a supplementary EPIC-KITCHENS block
under \emph{Full Assist}; strict-tolerance quality again improved,
while workload increased because the clip was longer and semantically
more complex. Full details are reported in the appendix.

\subsection{Full-Dataset Offline Mechanism Analysis on IMPACT}
\label{subsec:eval_mechanism}

\begin{figure}[t]
    \centering
    \includegraphics[width=\columnwidth]{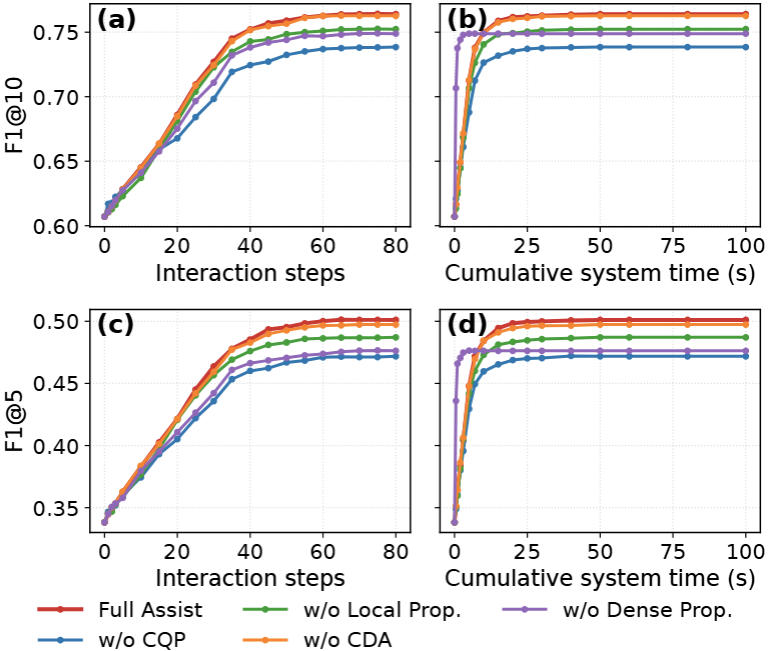}
    \vskip-1ex
    \caption{\textbf{Budget curves under controlled offline diagnostics
    on IMPACT} (F1 $\uparrow$). Boundary F1 vs.\ interaction steps
    (left) and cumulative system time (right) for F1@10 (top) and
    F1@5 (bottom).}
    \label{fig:budget_curves}
    \vskip-2ex
\end{figure}

Using the oracle-controlled offline protocol defined above,
Figure~\ref{fig:budget_curves} shows that removing any single component degrades boundary quality per interaction step relative to the full system across all 135 IMPACT test cases.

\noindent\textbf{Ablation.}
Table~\ref{tab:ablation} reports component ablations under a fixed interaction budget. Local proposal ablations test whether gains come from learned local correction inference and consistency-aware training; system-level ablations test whether query planning, adaptation, and structured propagation improve the correction loop over time. The oracle answerer supplies the ground-truth boundary location (snapped within $\tau{=}5$\,frames) and corresponding side labels; the resulting scores are upper-bound process diagnostics and are not directly comparable to Table~\ref{tab:human_main}.

\begin{table}[t]
    \centering
    \caption{\textbf{Full-dataset offline component ablation on IMPACT}
    under a fixed interaction budget.}
    \label{tab:ablation}
    \setlength{\tabcolsep}{4pt}
    \renewcommand{\arraystretch}{1.08}
    \begin{tabular}{@{}lcccc@{}}
        \toprule
        \textbf{Variant}
        & \textbf{F1@5 $\uparrow$}
        & \textbf{F1@10 $\uparrow$}
        & \textbf{F1@25 $\uparrow$}
        & \textbf{Edit $\uparrow$} \\
        \midrule
        Full Assist
            & \textbf{0.501} & \textbf{0.764} & 0.975 & 0.968 \\
        \hspace{0.4em}w/o CQP        & 0.472 & 0.739 & 0.973 & 0.972 \\
        \hspace{0.4em}w/o Local Prop.& 0.487 & 0.752 & 0.975 & 0.967 \\
        \hspace{0.4em}w/o CDA        & 0.497 & 0.763 & 0.977 & 0.972 \\
        \hspace{0.4em}w/o Dense Prop.& 0.476 & 0.749 & \textbf{0.985} & \textbf{0.990} \\
        \bottomrule
    \end{tabular}
    \vskip-4ex
\end{table}

Removing CQP or the local proposal module reduces strict-tolerance
quality relative to the full system, indicating that both query
selection and learned local correction matter for the final
trajectory. By contrast, disabling CDA changes the results only
modestly in this oracle-controlled setting, suggesting that most of
the gain already comes from the base interaction design and planner
backbone, while adaptation acts as a secondary refinement. Dense
propagation shows the opposite pattern: without it, looser-tolerance
F1 and Edit remain high, but strict boundary precision degrades,
consistent with propagation mainly improving local temporal
consistency around accepted corrections.

Session-level acceptance rates from the live study additionally show
that \emph{Full Assist} exhibits higher overall acceptance than
\emph{Scribble Only}, consistent with planner queries surfacing
boundary candidates that annotators are willing to confirm. The gap
between oracle scores and the live human-study results further
indicates that substantial room remains for translating component
capacity into user-facing gains.

\subsection{System Efficiency and Qualitative Analysis}
\label{subsec:eval_system}

IMPACT-Scribe treats sub-second system response as a design constraint.
Table~\ref{tab:latency_breakdown} reports system-side latency per
interaction (excluding human thinking and cursor motion), measured on
a laptop workstation (Nvidia RTX~4060 8\,GB), batch size 1, pre-cached I3D features.
The 95th-percentile total latency is 713\,ms ($p99 = 915$\,ms),
remaining below one second.

\begin{table}[t]
    \centering
    \caption{\textbf{System-side latency breakdown per interaction}
    ($\downarrow$).}
    \label{tab:latency_breakdown}
    \setlength{\tabcolsep}{4.5pt}
    \renewcommand{\arraystretch}{1.10}
    \begin{tabular}{lcc}
        \toprule
        \textbf{Component}
        & \textbf{Mean (ms)}
        & \textbf{Std (ms)} \\
        \midrule
        Feature lookup          & 19.9  & 16.8  \\
        Local proposal model    & 16.3  & 29.4  \\
        Query scoring           & 1.1  & 0.4  \\
        Dense propagation       & 282.5 & 147.2 \\
        Total system latency    & 319.7 & 164.9 \\
        \bottomrule
    \end{tabular}
    \vskip-3ex
\end{table}

\begin{figure}[t]
    \centering
    \includegraphics[width=\columnwidth,trim={0 1.2cm 0 0},clip]{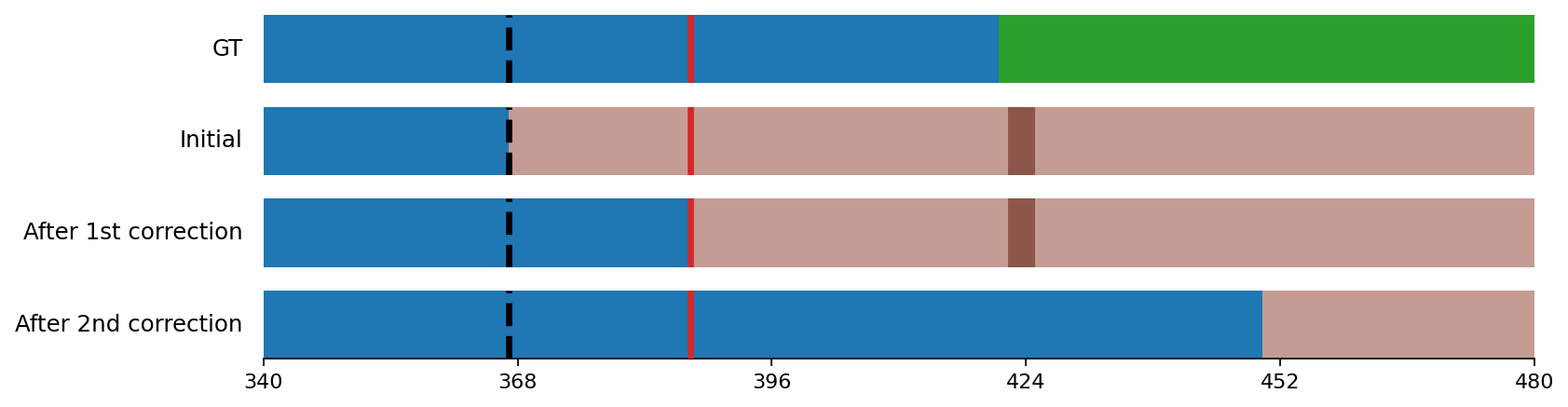}
    \caption{\textbf{Repeated correction with structured propagation.}}
    \label{fig:qualitative_b}
    \vskip-4ex
\end{figure}

Qualitative examples illustrate the system's operational behavior. Figure~\ref{fig:qualitative_b} shows a repeated-correction case in which the decoder re-optimizes the intervening uncertain span under soft anchor-consistency terms while preserving confirmed boundary cuts during writeback, restoring global coherence without manual repair of the full conflict region. In this cases, IMPACT-Scribe preserves annotator authority through explicit per-step acceptance while amplifying the downstream effect of each human decision—the core principle of correction-driven human--machine collaboration.


%% file: sec/conclusion.tex
\section{Conclusion}

Dense temporal annotation has long been treated as a throughput problem. IMPACT-Scribe reframes it as a collaboration problem: the more effectively a system can learn from each correction, the better it can allocate future assistance and convert human effort into annotation quality. By treating every accepted edit simultaneously as an update to the current labeling, a constraint for downstream propagation, and supervision for future query planning and proposal behavior, IMPACT-Scribe establishes a correction loop that improves over time rather than remaining static. Our human study and controlled offline analysis show that this closed-loop design yields stronger strict-boundary quality and higher dense-label quality per unit of annotator effort, while remaining practical for real interactive use. More broadly, we believe the same principle extends beyond temporal action annotation to other dense sequential labeling tasks in which human effort remains the primary bottleneck.